\documentclass[preprint,12pt]{elsarticle}
\makeatletter
\def\ps@pprintTitle{%
 \let\@oddhead\@empty
 \let\@evenhead\@empty
 \def\@oddfoot{}%
 \let\@evenfoot\@oddfoot}
\makeatother

\usepackage{graphicx}
\usepackage{amssymb}

\usepackage{lineno}

\usepackage{subfigure}

\begin{document}

\begin{frontmatter}


\title{A Formal Evaluation of PSNR as Quality Measurement Parameter for Image Segmentation Algorithms}



\author{Fernando A. Fardo}
\author{Victor H. Conforto}
\author{Francisco C. de Oliveira}
\author{Paulo S. Rodrigues}

\address{Centro Universit\'ario da FEI, S\~ao Paulo, Brazil}

\begin{abstract}

Quality evaluation of image segmentation algorithms are still subject of debate and research. Currently, there is no generic metric that could be applied to any algorithm reliably. This article contains an evaluation for the PSRN (Peak Signal-To-Noise Ratio) as a metric which has been used to evaluate threshold level selection as well as the number of thresholds in the case of multi-level segmentation. The results obtained in this study suggest that the PSNR is not an adequate quality measurement for segmentation algorithms.

\end{abstract}


\end{frontmatter}

\section{Introduction}
\label{sec:introduction}

In image processing, segmentation is a a set of techniques that separate regions from a scene based on similarity. There are several techniques available for this process \cite{Rodrigues2011,Erdmann2015}. Segmentation is usually based on attributes such as color, brightness contrast or continuity of pixel regions. In the particular case of threshold based techniques, one ore more threshold values is determined. Pixels of similar brightness levels are then grouped as below or above such threshold levels \cite{gonzalez2002digital}. 

\begin{figure}[h!]
  \centering
  \includegraphics[width=5cm]{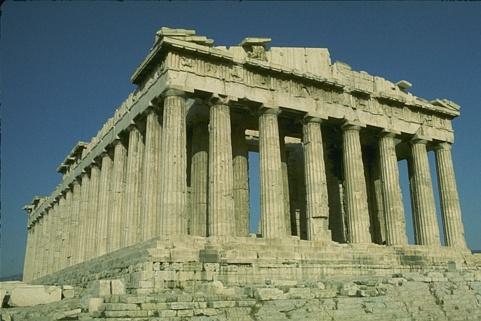}
  \caption{Example of an image with foreground and background}
   \label{f:imgTemple}
\end{figure}

\begin{figure}[h!]
\centering
\includegraphics[width=5cm]{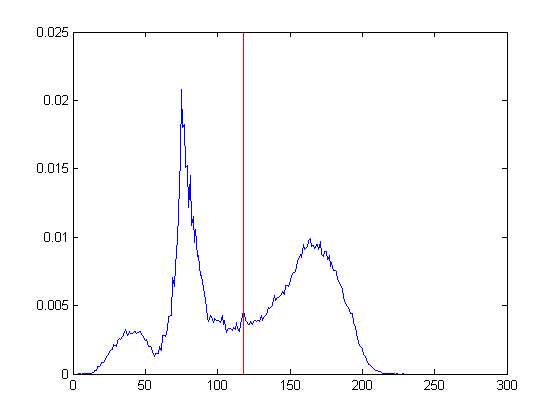}
\caption{Gray level histogram with detected threshold $t=118$}
 \label{f:imgHist}
\end{figure}

\begin{figure}[h!]
\centering
\includegraphics[width=5cm]{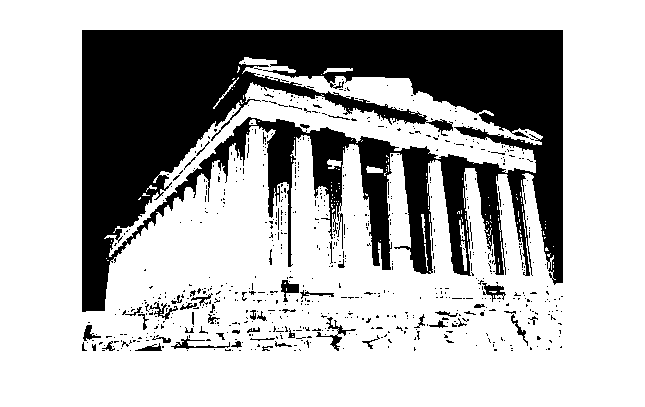}
\caption{Resulting image after threshold based segmentation with $t=118$}
 \label{f:imgBw}
\end{figure}

Fig. \ref{f:imgTemple} shows an example of a scene containing a simple foreground and a background. Fig. \ref{f:imgHist} shows it's corresponding $256$ gray level histogram with an obtained threshold level $t$ at $118$. The resulting image of a threshold based segmentation algorithm can is shown at Fig. \ref{f:imgBw}, where pixels below $t$ are set to $(0)$. Conversely, pixels of brightness level above $t$ are set to $255$. In this case, pixels labeled as (0) and (255) can be treated as the background and foreground, respectively.

Such techniques are often used at pre-processing step in high level computer vision based systems as it reduces the amount of irrelevant information by similarity grouping of the pixels in  the same region. The objective of threshold algorithms  is to detect the threshold level that separates an image in regions of interest more accurately. The main problem is that the quality evaluation of such algorithms lacks an objective parameter and cannot be determined automatically.

There are many proposals for a generic metric of segmentation algorithms. Such metric is often difficult to describe making an objective evaluation method potentially unreliable. The evaluation methods can be divided in two main categories: analytic and empirical \cite{cardoso2005toward}. The analytic methods are based in properties obtained from the segmented image which can be used in order to obtain a quantitative quality measurement. These methods are not very reliable as determining the quality of a segmentation based purely in analytic parameters can be difficult \cite{cardoso2005toward}. The empirical methods are based on the comparison of the resulting segmented image with pre-defined desirable results determined by human operators, and can be further divided into two subcategories, goodness methods and discrepancy methods. Goodness methods are uses pre-established parameters such as as region uniformity or inter region contrast. The discrepancy methods rely on the comparison of the segmentation result with a reference image known as ground truth, which is established by an human operator \cite{cardoso2005toward}.

Despite it's limitations, the PSNR has been used as an analytic metric by several authors of threshold based algorithms. \cite{chen2011gray,horng2011multilevel,arora2008multilevel}. As subject to study we performed some experiments to verify if PSNR can be used reliably as an analytic metric for image segmentation.

\section{PSNR}

The PSNR is a signal processing measurement that compares a given received or processed  signal to it's original source signal. This comparison allows us to quantify how much a processed signal is faithful to the original, also allowing us to identify possible noises or distortions to the signal. We can say that the PSNR represents a direct relationship of a signal before and after a degradation process.

Mathematically, the PSRN is described by the Equations (\ref{eq:psnr}) and (\ref{eq:mse})

\begin{equation}
\label{eq:psnr}
PSNR = 20  \log_{10}\left ( \frac{MAX_{I}^{2}}{\sqrt{MSE}} \right )
\end{equation}

\begin{equation}
\label{eq:mse}
MSE = \frac{1}{m n}\sum_{i=0}^{m-1}\sum_{j=0}^{n-1}\left [ I\left ( i,j \right ) - K\left ( i,j \right )\right ]^{2}
\end{equation}

where $MAX$ is the highest possible value of the signal. In the case of a gray scale image of 8 bits, $MAX=255$. As demonstrated in Eq. (\ref{eq:psnr}), the $PSNR$ is inversely proportional to the MSE (Mean Squared Error). The final value of the $PSNR$ is given in decibel.

The PSNR is generally used to evaluate the quality if transmission and compression of image or video signals, based on de mean square error of the received or processed image in comparison to the source image. However, it also has been used as an analytic metric for segmentation algorithm evaluation \cite{chen2011gray,horng2011multilevel}. In the case of multi-threshold algorithms, it was also used as a metric to determine the number of thresholds  \cite{arora2008multilevel} as well as it's values \cite{yun2011multi}.

\section{Objective}

The purpose of this paper is to evaluate the PSNR itself as a reliable analytic method for evaluation of image segmentation algorithms. 

\section{Methodology}

Since we are not trying to evaluate an algorithm but the metric itself, we cannot rely on some existing study that used the PSNR as an analytic method for evaluation. Instead, we propose the adoption of ground truth data that would normally be suitable for empiric methods as results of a segmentation algorithm. Then, we use the PSRN as an analytic method to evaluate such results.

\begin{figure}[h!]
\center
    \subfigure[ref1][]{\includegraphics[width=3cm]{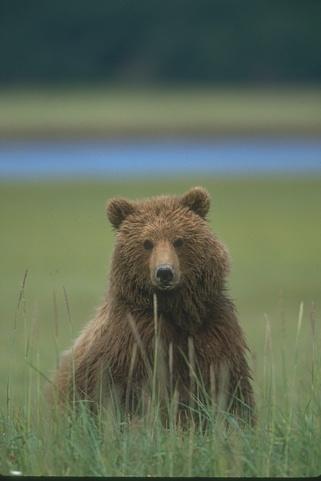}\label{f:imgOrigina}}
    \subfigure[ref2][]{\includegraphics[width=3cm]{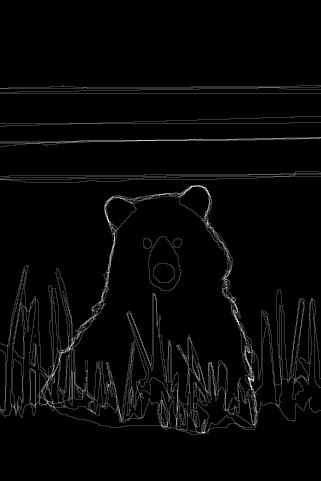}\label{f:imgOuro}}
    \caption{Example of an image from the database (a) and it's respective ground truth (b)}
    \label{f:berkeley}
\end{figure}

For the experiments, we used the set of images from the Berkeley BSR300 Database \cite{MartinFTM01}. It comprises of 300 images containing several types of scenes where every image $I$ has it's corresponding ground truth image $G$. The ground truth is an image contained contours of objects from each scene defined by volunteers as the most relevant ones. Fig. \ref{f:berkeley} shows an example of an image (a) of the database and it's respective ground truth image  (b).

\begin{figure}[h!]
\center
    \subfigure[ref1][]{\includegraphics[width=5cm]{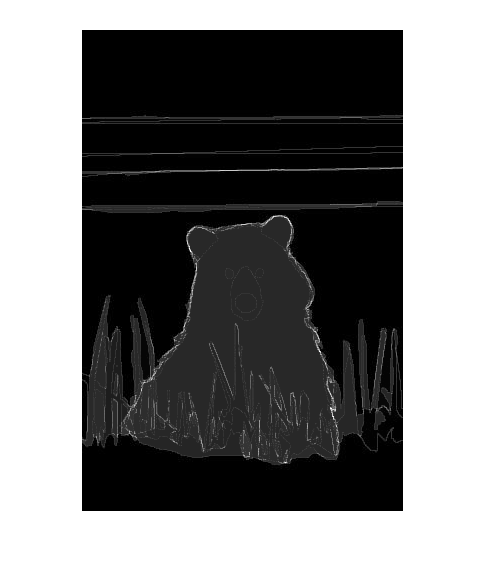}\label{f:imgFilled}}
    \subfigure[ref2][]{\includegraphics[width=5cm]{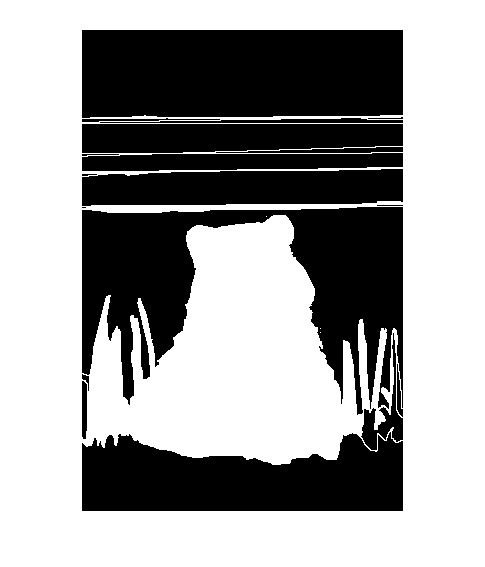}\label{f:imgSeg1}}
    \caption{Automatically filled ground truth image (a) and obtained binary mask (b) }
    \label{f:berkeleySegmentado}
\end{figure}

From each ground truth image $G$, a region mask $G'$ is obtained, separating the background from the foreground. The mask was obtained by automatically filling of the closed contours with the white color ($255$), thus creating masks with the most relevant regions of interest. After applying a threshold algorithm to this mask, a binary mask $B$ is obtained. Since computer vision techniques are strongly inspired by the human vision, we can assume that such binary masks are close to an ideal segmentation algorithm. Fig. \ref{f:berkeleySegmentado} shows an example of a filled ground truth $G'$ (a) and the corresponding binary mask $B$ (b) after threshold.

\begin{figure}[h!]
\center
    \subfigure[ref1][]{\includegraphics[width=5cm]{./segmentado.png}\label{f:imgSeg2}}
    \subfigure[ref2][]{\includegraphics[width=5cm]{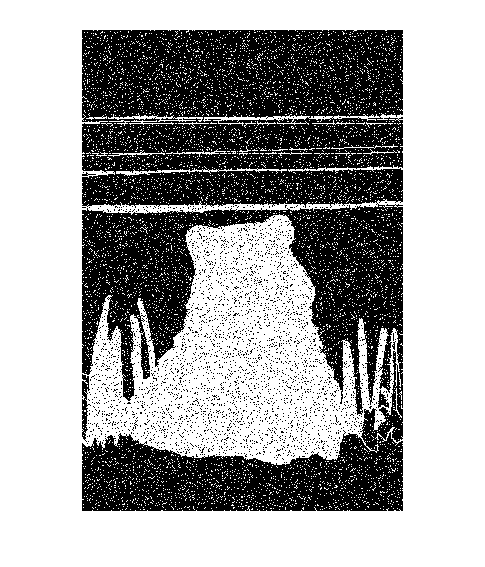}\label{f:imgSegNoised}}
    \caption{Binary mask $B$ (a) and bad segmentation mask $B'$ after salt and pepper noise (b) }
    \label{f:berkeleySalt}
\end{figure}

To verify the efficacy of the PSNR as an analytic method for image segmentation, we generated poorly segmented masks based on binary masks with the use of salt and pepper noise. As the salt and pepper noise adds changes pixels randomly to either $0$ or $255$ we can use this to simulate a bad segmentation. The resulting mask $B'$ therefore, contains several pixels that are incorrectly classified as foreground ($255$) and background ($0$). Fig. \ref{f:berkeleySalt} shows an example of a binary mask $B$ (a) and it's corresponding bad segmentation $B'$ (b).

When used as an analytic method, the PSNR is used between the resulting image and the original. Therefore, the PSNR must be calculated between each original image $I$ and the corresponding segmentation mask $B$ and bad segmentation mask $B'$.

For each image in the database, the PSNR is calculated between both $B$ and $B'$ and $I$ and the results of the PSNR are calculated and stored for posterior analysis.

\subsection{Proof}

Let $P$ be the set of PSNR results calculated between each binary mask $B$ and it's corresponding image $I$. Le $P'$ be the set of PSNR results calculated between each bad segmentation mask $B'$ and it's corresponding source image $I$.
If the PSNR is not an adequate analytic method, the average of PSNR values in $P$ should be significantly superior to those obtained in $P'$. For this paper, this condition is adopted as our main hypothesis.

\section{Results and discussion}

\begin{figure}[h!]
\centering
\includegraphics[width=5cm]{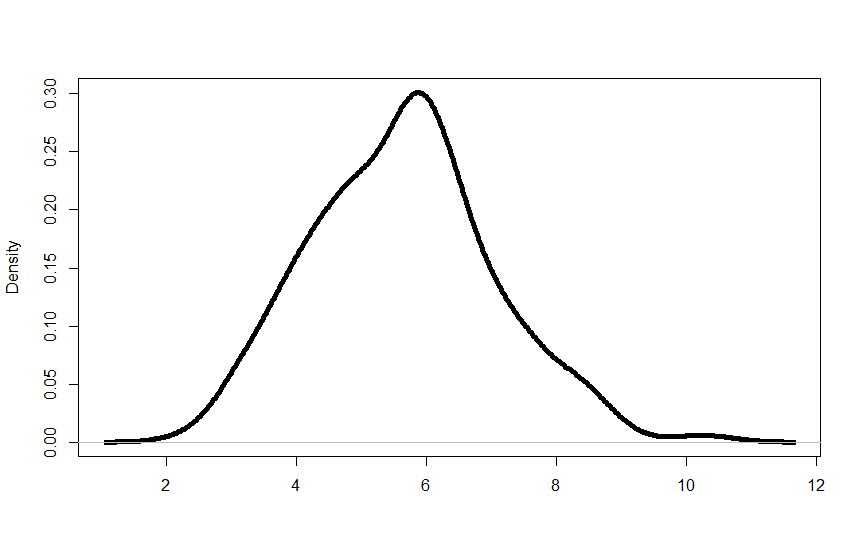}
\caption{Probability density for the set $P$ of PSNR results for good segmentation masks}
 \label{f:imgNormHuman}
\end{figure}

\begin{figure}[h!]
\centering
\includegraphics[width=5cm]{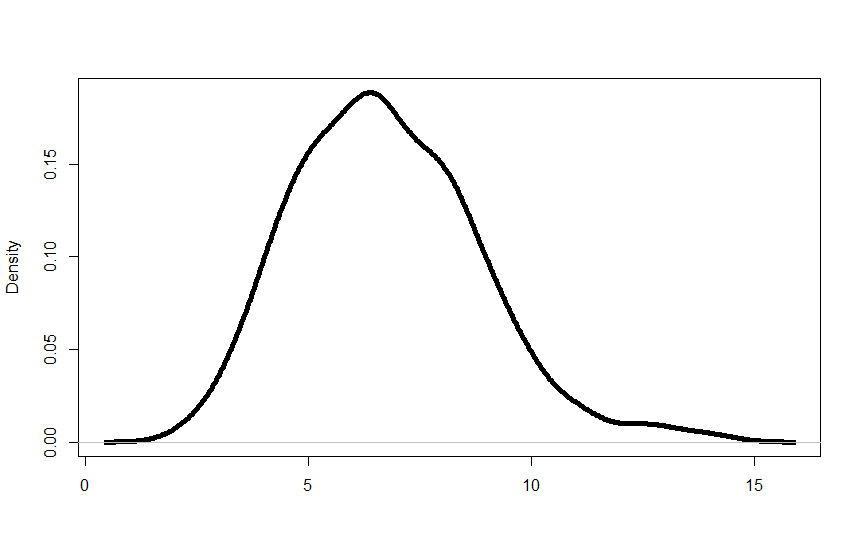}
\caption{Probability density for the set $P'$ of PSNR results for bad segmentation masks}
 \label{f:imgNormNoise}
\end{figure}

To confirm the main hypothesis, initially we proposed the use of Sudent's T test with $95\%$ of significance  \cite{student1908probable} between $P$ and $P'$. However this test requires the variance between the samples to be homogeneous. Firstly we used the Fisher's F test for variance \cite{fisher1941asymptotic} to verify such homogeneity between $P$ and $P'$. Figs. \ref{f:imgNormHuman} and \ref{f:imgNormNoise} shows the density of probability for the sets $P$ and $P'$ respectively. If the results from the F test indicate that the variance between the sets $P$ and $P'$ is not homogeneous the Student's T test cannot be applied. In this case, the Welch's T test should be used instead \cite{welch1947generalization}. These hypothesis tests were performed using the R language.

\subsection{Fisher's F test for variance}

As a null hypothesis for the F test, we adopt that the variances of the sets are homogeneous. As the alternative hypothesis, we adopt that the variances between the sets are not homogeneous. The results from the F test are shown on table \ref{tab:testF}.

\begin{table}[h!]
\centering
    \begin{tabular}{ | l | l | }
      \hline
      F&  0.4618  	\\ \hline
      df&  299\\ \hline
      df denominator& 299 \\ \hline
      P value & $4.265 {10}^{-11}$    \\ \hline
      Confidence interval & 0.3679506 a 0.5795227   \\ \hline
      Variance rates&  $0.4617745$   \\ \hline
      \end{tabular}
     \caption{Results for the F test of variance between $P$ and $P'$}
   \label{tab:testF}
\end{table}

The $p$ value for the F test is in the region for acceptance of the alternative hypothesis. Therefore, is not safe to assume that the variances between $P$ and $P'$ are homogeneous and the Student's T test cannot be used reliably. The Welch's T test is then used to determine if the difference between $P$ and $P'$ is statistically significant.

\subsection{Welch's T test}

As a null hypothesis, we adopt that $P$ and $P'$ are equal and the difference between the means of both sets is zero ($0$). As the alternative hypothesis, we adopt that the mean of $P'$ is superior to the mean of $P$. Should the alternative hypothesis be accepted, it would suggest that the bad segmentation masks were considered better then the ideal segmentation according to the PSNR metric.

The  Welch's T Test is then applied  with $95\%$ of significance between both sets $P$ and $P'$.  Table \ref{tab:testT} shows the results of the Welch's T test.

\begin{table}[h!]
\centering
    \begin{tabular}{ | l | l | }
      \hline
      T statistics&  -7.6524  	\\ \hline
      df&  526.607\\ \hline
      p value& $4.735 {10}^{-14}$    \\ \hline
      Confidence interval& $-\infty -0.8641351$   \\ \hline
      Mean of $P$& $5.638749$  \\ \hline
      Mean of $P'$& $6.740013$   \\ \hline
      \end{tabular}
     \caption{Results for the Welch's T test between $P$ and $P'$}
   \label{tab:testT}
\end{table}

The $p$ value for the Welch's T test is $4.735 \cdot {10}^{-14 }$ and is found in the area of rejection of the null hypothesis. We are left with the acceptance of the alternative hypothesis which indicate that the PSNR values calculated from the bad segmentation masks $B'$ are superior to the ones calculated by human obtained masks $B$.

\section{Final considerations}

We investigated the efficacy of the PSNR as an analytic method for segmentation algorithms the same way it's adopted. We used human created segmentation masks as an ideal reference of a segmentation algorithm and compared the calculated PSNR values from these masks to those calculated from artificially inferior segmentation masks.

To verify if the PSNR is a good evaluation method we compared the values of two sets of calculated PSNR values from good and bad segmentation masks. The mask generation procedure can produce masks that would not be obtainable from threshold algorithms as the values for labels are usually determined by the values of the calculated thresholds. For example, a foreground object on a brighter background would have it's pixels set to ($0$) in the binary mask while the background would be set to ($255$). However, there is no rule for what levels each label should be set to and this could influence the PSNR as well. Some graph based algorithms even separate regions using random colors \cite{huang2012robust}. Results from such such algorithms could not be verified with the PSNR as it is as they would change greatly from one execution to another.

We proposed the use of Welch's T test to verify if the difference between the sets of PSNR values from good and bad segmentation is significant. Higher PSNR values for good segmentation masks would suggest the PSNR is in fact a good analytic method. However, the results from the Welch T test suggest exactly the opposite. The values of PSNR value for the bad segmentation masks are significantly superior than the ones for good segmentation masks.  Therefore, the PSNR should not be considered an adequate method for evaluation of segmentation algorithms. However, the PSNR is still a good method to evaluate discrepancies between images and could  be used to evaluate edge detection algorithms by comparing with ground truth images such as the ones present in the BSR300 database. 

Future works could include the verification of multi-threshold algorithms and the determination of the number of thresholds as well as the impact of the label values.

\section{Acknowledgment}

The authors would like to thank the Berkeley University for the creation and availability of the BSR300 database.

\bibliography{referencias} \bibliographystyle{plain}

\begin{thebibliography}{10}

\bibitem{arora2008multilevel}
Siddharth Arora, Jayadev Acharya, Amit Verma, and Prasanta~K Panigrahi.
\newblock Multilevel thresholding for image segmentation through a fast
  statistical recursive algorithm.
\newblock {\em Pattern Recognition Letters}, 29(2):119--125, 2008.

\bibitem{cardoso2005toward}
Jaime~S Cardoso and Lu{\'\i}s Corte-Real.
\newblock Toward a generic evaluation of image segmentation.
\newblock {\em Image Processing, IEEE Transactions on}, 14(11):1773--1782,
  2005.

\bibitem{chen2011gray}
Yu-Kumg Chen, Fan-Chieh Cheng, and Pohsiang Tsai.
\newblock A gray-level clustering reduction algorithm with the least< i>
  psnr</i>.
\newblock {\em Expert Systems with Applications}, 38(8):10183--10187, 2011.

\bibitem{Erdmann2015}
H.~Erdmann, G.~Wachs-Lopes, C.~Gall{\~a}o, P.~M. Ribeiro, and S.~P. Rodrigues.
\newblock {\em Developments in Medical Image Processing and Computational
  Vision}, chapter A Study of a Firefly Meta-Heuristics for Multithreshold
  Image Segmentation, pages 279--295.
\newblock Springer International Publishing, Cham, 2015.

\bibitem{fisher1941asymptotic}
Ronald~Aylmer Fisher.
\newblock The asymptotic approach to behrens's integral, with further tables
  for the d test of significance.
\newblock {\em Annals of Eugenics}, 11(1):141--172, 1941.

\bibitem{gonzalez2002digital}
Rafael~C Gonzalez and Richard~E Woods.
\newblock Digital image processing, 2002.

\bibitem{horng2011multilevel}
Ming-Huwi Horng and Ren-Jean Liou.
\newblock Multilevel minimum cross entropy threshold selection based on the
  firefly algorithm.
\newblock {\em Expert Systems with Applications}, 38(12):14805--14811, 2011.

\bibitem{huang2012robust}
Qing-Hua Huang, Su-Ying Lee, Long-Zhong Liu, Min-Hua Lu, Lian-Wen Jin, and
  An-Hua Li.
\newblock A robust graph-based segmentation method for breast tumors in
  ultrasound images.
\newblock {\em Ultrasonics}, 52(2):266--275, 2012.

\bibitem{MartinFTM01}
D.~Martin, C.~Fowlkes, D.~Tal, and J.~Malik.
\newblock A database of human segmented natural images and its application to
  evaluating segmentation algorithms and measuring ecological statistics.
\newblock In {\em Proc. 8th Int'l Conf. Computer Vision}, volume~2, pages
  416--423, July 2001.

\bibitem{Rodrigues2011}
Paulo~S. Rodrigues and Gilson~A. Giraldi.
\newblock Improving the non-extensive medical image segmentation based on
  tsallis entropy.
\newblock {\em Pattern Analysis and Applications}, 14(4):369--379, 2011.

\bibitem{student1908probable}
Student.
\newblock The probable error of a mean.
\newblock {\em Biometrika}, pages 1--25, 1908.

\bibitem{welch1947generalization}
Bernard~L Welch.
\newblock The generalization ofstudent's' problem when several different
  population variances are involved.
\newblock {\em Biometrika}, pages 28--35, 1947.

\bibitem{yun2011multi}
Cao Yun-Fei, Xiao Yong-Hao, Yu~Wei-Yu, and Chen Yong-Chang.
\newblock Multi-level threshold image segmentation based on psnr using
  artificial bee colony algorithm.
\newblock {\em China Research Journal of Applied Sciences, Engineering and
  Technology Published: January}, 15, 2011.

\end{thebibliography}

\end{document}